\documentclass[10pt,journal,compsoc]{IEEEtran}
\usepackage{pifont}

\usepackage{ifpdf}

\ifCLASSOPTIONcompsoc
\usepackage[nocompress]{cite}
\else
\usepackage{cite}
\fi
\ifCLASSINFOpdf
\usepackage[pdftex]{graphicx}
\else
\usepackage[dvips]{graphicx}
\fi
\usepackage{amsmath}
\usepackage{algorithmic}

\usepackage{array}

\ifCLASSOPTIONcompsoc
\usepackage[caption=false,font=footnotesize,labelfont=sf,textfont=sf]{subfig}
\else
\usepackage[caption=false,font=footnotesize]{subfig}
\fi
\usepackage{url}

\usepackage{color}
\usepackage{lipsum}
\usepackage{algorithm}

\usepackage{graphicx}
\usepackage{multirow}
\usepackage{multicol}
\usepackage{pifont}
\usepackage{wrapfig}
\usepackage{amsmath} 
\usepackage{amssymb}
\usepackage{subfig}
\usepackage{booktabs}
\usepackage{ragged2e}
\usepackage[marginal]{footmisc}
\newtheorem{theorem}{Theorem}
\newcommand{\minisection}[1]{\vspace{.02in}\noindent{\textbf{#1}}.}

\usepackage{algorithm}
\usepackage{algpseudocode}

\begin{document}
	\title{Generalized Kullback-Leibler Divergence Loss}
	\author{Jiequan~Cui,
                Beier~Zhu,
                Qingshan~Xu,
		    Zhuotao~Tian,
                Xiaojuan~Qi,
                Bei~Yu,
                Hanwang~Zhang,
                Richang Hong
		
		\IEEEcompsocitemizethanks{\IEEEcompsocthanksitem J.~Cui and R.~Hong are with Hefei University of Technology (HFUT). B.~Zhu and Q.~Xu are with the University of Science and Technology of China (USTC). H.~Zhang is with Nanyang Technological University (NTU), Singapore. B.~Yu is with The Chinese University of Hong Kong (CUHK), Sha Tin, Hong Kong. X.~Qi is with The University of Hong Kong (HKU). Z.~Tian is with Harbin Institute of Technology, Shenzhen. \protect\\
			E-mail: jiequancui@gmail.com }}

    \IEEEtitleabstractindextext{%
\begin{abstract}
\justifying 
In this paper, we delve deeper into the Kullback–Leibler (KL) Divergence loss and mathematically prove that it is equivalent to the Decoupled Kullback-Leibler (DKL) Divergence loss that consists of 1) a weighted Mean Square Error ($\mathbf{w}$MSE) loss and 2) a Cross-Entropy loss incorporating soft labels. 
Thanks to the decoupled structure of DKL loss, we have identified two areas for improvement.
Firstly, we address the limitation of KL loss in scenarios like knowledge distillation by breaking its asymmetric optimization property along with a smoother weight function. This modification effectively alleviates convergence challenges in optimization, particularly for classes with high teacher-predicted scores.
Secondly, we introduce class-wise global information into KL/DKL to reduce bias arising from individual samples.
With these two enhancements, we derive the Generalized Kullback–Leibler (GKL) Divergence loss and evaluate its effectiveness by conducting experiments on CIFAR-10/100, ImageNet, and vision-language datasets, focusing on adversarial training and knowledge distillation tasks. Specifically, we achieve new state-of-the-art adversarial robustness on the public leaderboard --- \textit{RobustBench} and competitive knowledge distillation performance across CIFAR/ImageNet models and CLIP models, demonstrating the substantial practical merits. Our code is available at \url{https://github.com/jiequancui/DKL}.
\end{abstract}
}
	
\maketitle
\IEEEdisplaynontitleabstractindextext
\IEEEpeerreviewmaketitle
    
\IEEEraisesectionheading{\section{Introduction}\label{sec:intro}}

\IEEEPARstart{L}{oss} functions are a critical component of training deep models.
Cross-Entropy loss is particularly important in image classification tasks~\cite{he2016deep, DBLP:conf/cvpr/TanCPVSHL19, dosovitskiy2020image, cui2019fast}, while mean square error (MSE) loss is commonly used in regression tasks~\cite{ren2015faster, he2017mask, he2022masked}.
Contrastive loss~\cite{chen2020simple, DBLP:conf/cvpr/He0WXG20, DBLP:conf/nips/GrillSATRBDPGAP20, DBLP:conf/nips/CaronMMGBJ20, cui2021parametric, cui2022generalized} has emerged as a popular objective for representation learning. 
The selection of an appropriate loss function can exert a substantial influence on a model's performance. Therefore, the development of effective loss functions~\cite{cao2019learning, focalloss, dkd, wang2020improving, johnson2016perceptual, berman2018lovasz, wen2016discriminative, tan2020equalization, cui2022region, cui2021parametric} remains a critical research topic in the fields of computer vision and machine learning.

Kullback-Leibler (KL) Divergence quantifies the degree of dissimilarity between a probability distribution and a reference distribution.
As one of the most frequently used loss functions, it finds application in various scenarios, such as adversarial training~\cite{zhang2019theoretically, wu2020adversarial, cui2021learnable, Jia_2022_CVPR}, knowledge distillation~\cite{kd,revkd,dkd}, incremental learning~\cite{chaudhry2018riemannian, lee2017overcoming}, and robustness on out-of-distribution data~\cite{hendrycks2019augmix}.
Although many of these studies incorporate KL Divergence loss as part of their algorithms, they may not thoroughly investigate the underlying mechanisms of the loss function.
To bridge this gap, our paper aims to elucidate the working mechanism of KL Divergence regarding gradient optimization.

\begin{figure}[t]
    \centering
    \includegraphics[width=.95\linewidth]{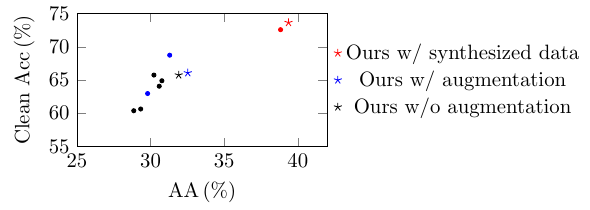}
    \vspace{-0.1in}
    \caption{\textbf{We achieve SOTA robustness on CIFAR-100.} ``star'' represents our method while ``circle'' denotes previous methods. ``Black'' means adversarial training with image preprocessing only including random crop and flip, ``Blue'' is for methods with AutoAug or CutMix, and ``Red'' represents methods using synthesized data. AA is short for Auto-Attack~\cite{croce2020reliable}.}
    \label{fig:cifar100_sota}
    \vspace{-0.1in}
\end{figure}

\begin{figure*}[tb!]
    \centering
    \includegraphics[width=.8\linewidth]{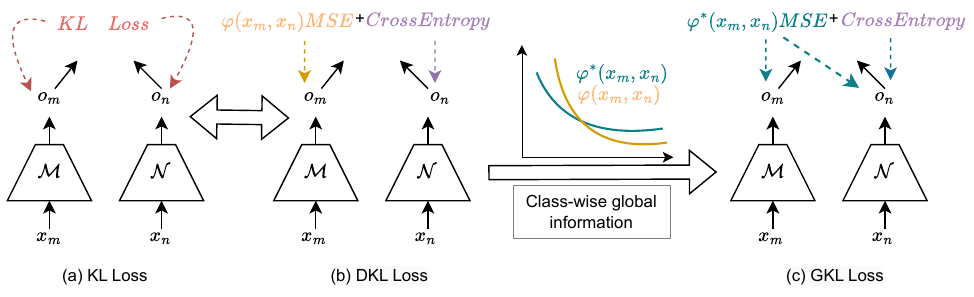}
    \vspace{-0.2in}
    \caption{\textbf{Comparisons of gradient backpropagation between KL, DKL, and GKL losses.} (b) DKL loss is equivalent to (a) KL loss regarding backward optimization.
    $\mathcal{M}$ and $\mathcal{N}$ can be the same one (like in adversarial training) or two separate (like in knowledge distillation) models determined by application scenarios. Similarly, $x_{m}$, $x_{n}$ $\in$ $X$ can also be the same one (like in knowledge distillation) or two different (like in adversarial training) images. $o_{m}$, $o_{n}$ are logits output with which the probability vectors are obtained when applying the \textit{softmax} activation. Solid arrows represent the forward process while dotted arrows indicate the backward process driven by the corresponding loss functions in the same color. $\varphi(x_{m},x_{n})$ is weight function depending on prediction of $x_{m}$. $\varphi^{*}(x_{m},x_{n})$ is our designed smoother weight function. It can be sample-wise or class-wise, determined by whether class-wise global information is incorporated. }
    \label{fig:dkl_ikl}
    \vspace{-0.1in}
\end{figure*}

\minisection{Deoupled Kullback-Leibler (DKL) Divergence Loss}
Our study focuses on the analysis of Kullback–Leibler (KL) Divergence loss from the perspective of gradient optimization. For models with \textit{softmax} activation, we provide theoretical proof that it is equivalent to the Decoupled Kullback–Leibler (DKL) Divergence loss, which comprises a weighted Mean Square Error ($\mathbf{w}$MSE) loss and a Cross-Entropy loss with soft labels. Figs.~\ref{fig:dkl_ikl}(a) and (b) reveal the equivalence between KL and DKL losses regarding gradient backpropagation. With the decoupled structure, it becomes more convenient to analyze how the KL loss works in training optimization.

\begin{figure}[tb!]
    \centering
    \subfloat[ImageNet-LT]            { \includegraphics[width=0.48\linewidth]{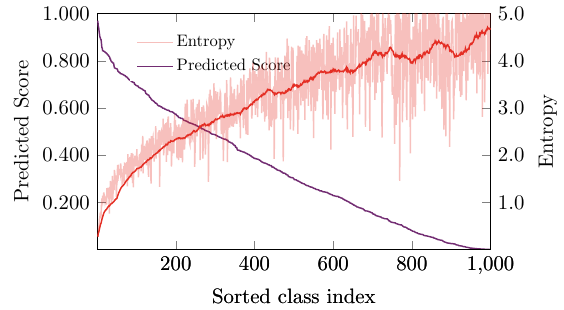} \label{fig:imagenetlt_score_entropy}}
    \subfloat[Full ImageNet]            { \includegraphics[width=0.48\linewidth]{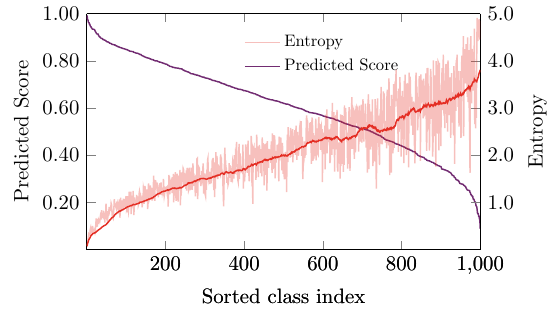}  \label{fig:imagenet_score_entropy}}
    \caption{
        \textbf{Classification models suffer from an imbalanced distribution of predicted scores.}
        (a) On ImageNet-LT;
        (b) On Full ImageNet;
        The higher the teacher-predicted score, the lower the entropy is expected to be for the student model during KD training.
    }
    \label{fig:score_entropy}
\end{figure}

\minisection{Generalized Kullback-Leibler (GKL) Divergence Loss}
We have identified potential issues of KL loss with the newly derived DKL loss.
Specifically, its gradient optimization is asymmetric regarding the inputs. As illustrated in Fig.~\ref{fig:dkl_ikl}(b), the gradients on $o_{m}$ and $o_{n}$ are asymmetric and driven by the $\mathbf{w}$MSE and Cross-Entropy, respectively. 
This optimization asymmetry can lead to the $\mathbf{w}$MSE component being ignored in certain scenarios, such as knowledge distillation where $o_{m}$ is the logits of the teacher model and often detached from gradient backpropagation.
Fortunately, it is convenient to break the asymmetric optimization property with the decoupled structure of DKL loss via enabling the gradient on $o_{n}$ from $\mathbf{w}$MSE as shown in Fig.~\ref{fig:dkl_ikl}(c).

In traditional knowledge distillation, we observe that teacher soft labels often exhibit an imbalanced distribution even if the model is trained on balanced datasets such as ImageNet. Distilling knowledge from such teachers with the KL loss can introduce optimization difficulties, especially for classes with high teacher prediction scores. As shown in Fig.~\ref{fig:score_entropy}, classes with higher teacher prediction scores require the student model to produce substantially lower-entropy outputs during KD training, which increases the convergence challenge. By breaking the asymmetric optimization property and adopting a properly designed smooth weighting function for the $\mathbf{w}$MSE component, the proposed DKL loss can effectively alleviate this issue.

Moreover, the $\mathbf{w}$MSE component is guided by sample-wise predictions, which can be noisy and unstable. In particular, hard examples or outliers with incorrect prediction scores can lead to challenging optimization. To address this issue, we incorporate class-wise global information to regularize the training process.
Integrating DKL with these two enhancements, we derive the Generalized Kullback–Leibler (GKL) Divergence loss.

\minisection{Our Results}
To demonstrate the effectiveness of the proposed GKL loss, we evaluate it on both adversarial training and knowledge distillation tasks. On CIFAR-10/100, GKL achieves new state-of-the-art robustness on the public \textit{RobustBench}\footnote{https://robustbench.github.io/} leaderboard. Comparisons with previous adversarial training methods are presented in Fig.~\ref{fig:cifar100_sota}. 
For knowledge distillation, in addition to experiments on ImageNet and CIFAR, we further evaluate GKL on CLIP models~\cite{cherti2023reproducible,radford2021learning} using vision-language data. The resulting performance gains are verified on zero-shot ImageNet classification and on the autoregressive vision-language model LLaVA~\cite{liu2023llava}.

In summary, the main contributions of our work are:
\begin{itemize} 
\item We reveal that the KL loss is mathematically equivalent to a composite of a weighted MSE ($\mathbf{w}$MSE) loss and a Cross-Entropy loss employing soft labels.
\item Based on our analysis, we propose two modifications for enhancement: breaking its asymmetric optimization and proper design of weight function $\varphi(x_{m}, x_{n})$ incorporating class-wise global information, deriving the Generalized Kullback–Leibler (GKL) loss.
\item With the proposed GKL loss, we obtain the state-of-the-art adversarial robustness on \textit{RobustBench} and competitive knowledge distillation performance on CIFAR-10/100, ImageNet, and CLIP models.
\end{itemize}

    \section{Related Work}
\label{sec:related_work}

\minisection{Adversarial Robustness}
Since the identification of adversarial examples by Szegedy et al.~\cite{szegedy2013intriguing}, the security of deep neural networks (DNNs) has gained significant attention,
and ensuring the reliability of DNNs has become a prominent topic in the machine learning community. Adversarial training~\cite{madry2017towards}, being the most effective method, stands out due to its consistently high performance.

Adversarial training incorporates adversarial examples into the training process.
Madary et al.~\cite{madry2017towards} propose the adoption of the universal first-order adversary, specifically the PGD attack, in adversarial training.
Zhang et al.~\cite{zhang2019theoretically} trade off the accuracy and robustness by the KL loss.
Wu et al.~\cite{wu2020adversarial} introduce adversarial weight perturbation to explicitly regulate the flatness of the weight loss landscape.
Cui et al.~\cite{cui2021learnable} leverage guidance from naturally-trained models to regularize the decision boundary in adversarial training.
Additionally, various other techniques~\cite{Jia_2022_CVPR} focusing on optimization or training aspects have also been developed. Besides, several recent works~\cite{gowal2021improving,wang2023better,addepalli2022efficient} have explored the use of data augmentation or synthesized data to improve adversarial training. We have explored the mechanism of KL loss for adversarial robustness in this paper. The effectiveness of the proposed GKL loss is tested in both settings with and without synthesized data~\cite{karras2022elucidating}.

\minisection{Knowledge Distillation}
The concept of Knowledge Distillation (KD) was first introduced by Hinton et al.~\cite{kd}.
It involves extracting ``dark knowledge'' from accurate teacher models to guide the learning process of student models. This is achieved by utilizing the KL loss to regularize the output probabilities of student models, aligning them with those of their teacher models when given the same inputs. This simple yet effective technique significantly improves the generalization ability of smaller models and finds extensive applications in various domains. Since the initial success of KD~\cite{kd}, several advanced methods, including logits-based~\cite{cho2019efficacy, furlanello2018born, mirzadeh2020improved, yang2019snapshot, zhang2018deep, dkd, huang2022knowledge} and features-based approaches~\cite{fitnets, crd, ofd, at, revkd, heo2019comprehensive, heo2019knowledge, kim2018paraphrasing, park2019relational, peng2019correlation, yim2017gift}, have been introduced. 
While prior works~\cite{DBLP:journals/corr/abs-2010-01809,xiang2020learning,li2022nested} have explored tackling the long-tailed problem with KD, the proposed DKL in this paper bridges the connection between the imbalanced learning and KD from a different perspective and could generalize to more general settings beyond the long-tailed problem.

\minisection{Other Applications of KL Divergence Loss}
In semi-supervised learning, the KL loss acts as a consistency loss between the outputs of weakly and strongly augmented images~\cite{sohn2020fixmatch, tarvainen2017mean}.
In continual learning, KL loss helps retain previous knowledge by encouraging consistency between the outputs of pre-trained and newly updated models~\cite{chaudhry2018riemannian, lee2017overcoming}.
Additionally, KL loss is also applied to enhance model robustness to out-of-distribution data~\cite{hendrycks2019augmix, zhu2023prompt,zhu2023debiased}.

\section{Method}
\label{sec:method}
In this section, we first introduce the preliminary mathematical notations in Sec.~\ref{sec:preliminary}. We then present the theoretical analysis of the equivalence between the KL and DKL losses in Sec.~\ref{sec:dkl}. Next, in Sec.~\ref{sec: IKL}, we propose the GKL loss to address potential limitations of KL/DKL. Finally, a case study with empirical verification is provided in Sec.~\ref{sec:case_study}.

\subsection{Preliminary}
\label{sec:preliminary}

\minisection{Definition of KL Divergence}
Kullback-Leibler (KL) divergence measures the difference between two probability distributions. For two continuous distributions $P$ and $Q$, it is defined as:
\begin{equation}
   D_{KL}(P||Q)=\int_{-\infty}^{+\infty} p(x)\log \frac{p(x)}{q(x)}\,dx,
   \label{eq_kl}
\end{equation}
where $p(x)$ and $q(x)$ denote the probability density functions of $P$ and $Q$, respectively.
The KL loss is one of the most widely used objectives in deep learning and is broadly applied in tasks involving categorical distributions. In this paper, we mainly study its role in adversarial training and knowledge distillation.

In adversarial training, the KL loss improves model robustness by aligning the output probability distribution of adversarial examples with that of their corresponding clean images, thus minimizing output changes despite input perturbations. In knowledge distillation, the KL loss enables a student model to mimic the behavior of a teacher model, facilitating knowledge transfer that enhances the student model’s generalization performance.

\minisection{Applications of KL Loss in Deep Learning}
We consider image classification models that output probability vectors through the \textit{softmax} activation. Let $\mathbf{o}_{i} \in \mathbb{R}^{C}$ denote the logits produced by a model for an input image $x_{i} \in X$, where $C$ is the number of classes. The corresponding predicted probability vector is $\mathbf{s}_{i} \in \mathbb{R}^{C}$, computed as $\mathbf{s}_{i}=\textit{softmax}(\mathbf{o}_{i})$. Here, $\mathbf{o}_{i}^{j}$ and $\mathbf{s}_{i}^{j}$ represent the logit and predicted probability of the $j$-th class, respectively.
The KL loss is commonly used to encourage similarity between two output distributions, $\mathbf{s}_{m}$ and $\mathbf{s}_{n}$, leading to the following objective:

\begin{equation}
   \mathcal{L}_{KL}(x_{m}, x_{n})=\sum_{j=1}^{C}\mathbf{s}_{m}^{j}\log\frac{\mathbf{s}_{m}^{j}}{\mathbf{s}_{n}^{j}},
   \label{eq_kl_sample}
\end{equation}

For example, in adversarial training, $x_{m}$ denotes a clean image and $x_{n}$ denotes its corresponding adversarial example. In knowledge distillation, $x_{m}$ and $x_{n}$ are the same image, but are separately fed into the teacher and student models. Notably, in knowledge distillation, $\mathbf{s}_{m}$ is detached from gradient backpropagation because the teacher model is pre-trained and kept fixed during training.

\subsection{Decoupled Kullback-Leibler Divergence Loss}
\label{sec:dkl}
Previous works~\cite{kd, zhang2019theoretically, cui2021learnable} incorporate the KL loss into their methods without explicitly exploring its inherent working mechanism. The goal of this paper is to uncover the driving force behind gradient-based optimization by examining the KL loss function. 
With the backpropagation rule in training optimization, the derivative gradients are as follows,
\begin{eqnarray}
     \frac{\partial \mathcal{L}_{KL}}{\partial \mathbf{o}_{m}^{j}} &=& \sum_{k=1}^{C} ((\Delta \mathbf{m}_{j,k} -\Delta \mathbf{n}_{j,k}) * (\mathbf{s}_{m}^{k} * \mathbf{s}_{m}^{j})), \label{eq_gradient_1} \\
     \frac{\partial \mathcal{L}_{KL}}{\partial \mathbf{o}_{n}^{j}} &=& \mathbf{s}_{n}^{j} - \mathbf{s}_{m}^{j},
     \label{eq_gradient_2}
\end{eqnarray}
where $\Delta \mathbf{m}_{j,k} = \mathbf{o}_{m}^{j} - \mathbf{o}_{m}^{k}$, and $\Delta \mathbf{n}_{j,k} = \mathbf{o}_{n}^{j} - \mathbf{o}_{n}^{k}$.

Leveraging the antiderivative technique alongside the structured gradient information, we introduce a novel formulation called the Decoupled Kullback-Leibler (DKL) Divergence loss, as presented in Theorem~\ref{thm:thm_dkl}. The DKL loss is designed to be equivalent to the KL loss while offering a more analytically tractable alternative for further exploration and study.

\begin{theorem}
	\label{thm:thm_dkl}
	\normalfont{From the perspective of gradient optimization, the Kullback-Leibler (KL) Divergence loss is equivalent to the following Decoupled Kullback-Leibler (DKL) Divergence loss when $\alpha=1$, $\beta =1$, and $\varphi(x_{m},x_{n})=\sqrt{\mathcal{S}(\mathbf{w}_{m})}$}:
        \begin{eqnarray}
            \mathcal{L}_{DKL}(x_m, x_n) =& \underbrace{\frac{\alpha}{4} \|\varphi(x_{m}, x_{n})(\Delta \mathbf{m} -\mathcal{S}(\Delta \mathbf{n}))\|^2}_{\textbf{weighted MSE (wMSE)}}  \nonumber \\
            &\underbrace{-\beta \cdot \mathcal{S}(\mathbf{s}_{m}^{\top}) \cdot \log \mathbf{s}_{n}}_{\textbf{Cross-Entropy}},
        \label{eq_dkl}
        \end{eqnarray}
        where $\mathcal{S}(\cdot)$ represents \textit{stop gradients} operation,
        $\mathbf{s}_{m}^{\top}$ is transpose of $\mathbf{s}_{m}$,
        $\mathbf{w}_{m}^{j,k}$ = $\mathbf{s}_{m}^{j} * \mathbf{s}_{m}^{k}$,
        $\Delta \mathbf{m}_{j,k} = \mathbf{o}_{m}^{j} - \mathbf{o}_{m}^{k}$, 
        and $\Delta \mathbf{n}_{j,k} = \mathbf{o}_{n}^{j} - \mathbf{o}_{n}^{k}$.
        Summation is used for the reduction of $\|\cdot\|^2$.

        \textit{Proof~~} 
        For KL loss, we have the following derivatives according to the chain rule:
\begin{eqnarray}
    \frac{\partial \mathbf{s}_{m}^{i}}{\partial \mathbf{o}_{m}^{i}} &=& \mathbf{s}_{m}^{i} * \sum_{j!=i}^{C} \mathbf{s}_{m}^{j},  \nonumber \\
    \frac{\partial \mathbf{s}_{m}^{j}}{\partial \mathbf{o}_{m}^{i}} &=& -\mathbf{s}_{m}^{i} * \mathbf{s}_{m}^{j}, \nonumber \\
    \frac{\partial \mathcal{L}_{KL}}{\partial \mathbf{s}_{m}^{i}} &=& \log \mathbf{s}_{m}^{i} - \log \mathbf{s}_{n}^{i} +1, \nonumber \\
    \frac{\partial \mathcal{L}_{KL}}{\partial \mathbf{o}_{n}^{i}} &=& \mathbf{s}_{n}^{i} - \mathbf{s}_{m}^{i} \label{kl_n} \\
    \frac{\partial \mathcal{L}_{KL}}{\partial \mathbf{o}_{m}^{i}} &=& \frac{\mathcal{L}_{KL}}{\partial \mathbf{s}_{m}^{i}} * \frac{\partial \mathbf{s}_{m}^{i}}{\partial \mathbf{o}_{m}^{i}} + \sum_{j!=i}^{C} \frac{\mathcal{L}_{KL}}{\partial \mathbf{s}_{m}^{j}} * \frac{\partial \mathbf{s}_{m}^{j}}{\partial \mathbf{o}_{m}^{i}} \nonumber \\
    &=& \sum_{j}^{C}(\Delta \mathbf{m}_{i,j} - \Delta \mathbf{n}_{i,j}) * \mathbf{w}_{m}^{i,j}
    \label{kl_m}
\end{eqnarray}

For DKL los, we expand the Eq.~\eqref{eq_dkl} as:
\begin{eqnarray}
            \mathcal{L}_{DKL}(x_m, x_n) \!=\!& \underbrace{\frac{\alpha}{4} \sum_{j=1}^{C} \sum_{k=1}^{C} (\Delta \mathbf{m}_{j,k} \!-\! \mathcal{S}(\Delta \mathbf{n}_{j,k}) )^2  
            \mathcal{S}(\mathbf{w}_{m}^{j,k})}_{\textbf{weighted MSE (wMSE)}} \nonumber \\
            &\underbrace{-\beta \sum_{j=1}^{C} \mathcal{S}(\mathbf{s}_{m}^{j}) \log \mathbf{s}_{n}^{j}}_{\textbf{Cross-Entropy }}, \nonumber
        \label{eq_dkl_expand}
\end{eqnarray}

According to the chain rule, we obtain the following equations:
\begin{eqnarray}
\frac{\partial \mathcal{L}_{DKL}}{\partial \mathbf{o}_{n}^{i}} &=& \beta * (\mathbf{s}_{n}^{i} - \mathbf{s}_{m}^{i})  \label{dkl_n}\\
\frac{\partial \mathcal{L}_{DKL}}{\partial \mathbf{o}_{m}^{i}} &=& \frac{\alpha}{4} * 2 * (\sum_{j}^{C} (\Delta \mathbf{m}_{j,i}-\Delta \mathbf{n}_{j,i}) * (-\mathbf{w}_{m}^{j,i}) \nonumber \\
&+& \sum_{k}^{C} (\Delta \mathbf{m}_{i,k} - \Delta \mathbf{n}_{i,k}) * \mathbf{w}_{m}^{i,k})\nonumber \\
&=& \alpha * \sum_{j}^{C}(\Delta \mathbf{m}_{i,j} - \Delta \mathbf{n}_{i,j}) * \mathbf{w}_{m}^{i,j}
\label{dkl_m}
\end{eqnarray}
Comparing Eq.~\eqref{kl_n} and Eq.~\eqref{dkl_n}, Eq.~\eqref{kl_m} and Eq.~\eqref{dkl_m}, we conclude that
DKL loss and KL loss have the same derivatives given the same inputs. Thus, KL loss is equivalent to DKL loss in terms of gradient optimization.
\end{theorem}

\minisection{Remark}
With Theorem~\ref{thm:thm_dkl}, we know that KL loss is equivalent to DKL loss regarding gradient optimization, \textit{i.e., DKL loss produces the same gradients as KL loss given the same inputs.} Therefore, KL loss can be interpreted as a composition of a $\mathbf{w}$MSE loss and a Cross-Entropy loss. This is the first work to reveal the accurate quantitative relationships between KL, Cross-Entropy, and MSE losses. Upon examining this new formulation, we identify two potential issues of KL loss.

\minisection{Asymmetirc Optimization Property}
As shown in Eqs.~\eqref{eq_gradient_1} and~\eqref{eq_gradient_2}, gradient optimization is asymmetric for $\mathbf{o}_{m}$ and $\mathbf{o}_{n}$.
The $\mathbf{w}$MSE and Cross-Entropy losses in Theorem~\ref{thm:thm_dkl} are complementary and collaboratively work together to make $\mathbf{o}_{m}$ and $\mathbf{o}_{n}$ similar. 
Nevertheless, the asymmetric optimization can cause the $\mathbf{w}$MSE component to be neglected or overlooked when $\mathbf{o}_{m}$ is detached from gradient backpropagation, which is the case for knowledge distillation, potentially leading to performance degradation.

We take knowledge distillation as an example to illustrate the necessity of the $\mathbf{w}$MSE component. As shown in Fig.~\ref{fig:score_entropy}, the model predicted scores are often imbalanced even when training it on balanced datasets such as ImageNet, a similar phenomenon also observed in previous work~\cite{cui2024classes}. 
In conventional knowledge distillation, the student is trained to match the teacher’s predicted score distribution. Consequently, classes with higher teacher prediction scores require the student outputs to attain much lower entropy than classes with lower teacher prediction scores, which can lead to more difficult optimization and slower convergence for those classes.
Fortunately, the $\mathbf{w}$MSE term, together with a properly designed weighting function $\varphi(x_{m}, x_{n})$, can effectively alleviate this issue, as discussed in Sec.~\ref{sec: IKL}.

\minisection{Sample-wise Prediction Bias}
As shown in Eq.~\eqref{eq_dkl}, the weighting function in the $\mathbf{w}$MSE component is defined as $\varphi(x_{m}, x_{n})=\sqrt{\mathbf{w}_{m}}$, which depends on the prediction score of the sample $x_{m}$. However, sample-wise predictions can exhibit large variance. In particular, incorrect predictions on hard examples or outliers can introduce misleading weights, resulting in unstable optimization. Our studies in Sections~\ref{sec:case_study} and~\ref{sec:ablation} further show that the choice of $\varphi(x_{m}, x_{n})$ has a significant impact on adversarial robustness.

\begin{table*}[tb!]
    \centering
    \caption{\textbf{Ablation study on weight function $\varphi(x_{m}, x_{n})$ and ``BA'' with DKL loss.} ``BA'' indicates ``Breaking Asymmetric Optimization''. ``Clean'' is the test accuracy of clean images and ``AA'' is the robustness under Auto-Attack. CIFAR-100 is used for the adversarial training task and ImageNet is adopted for the knowledge distillation task.}
    \vspace{-0.1in}
    {
    \begin{tabular}{lcccccc}
    \toprule
         \textbf{Index}    &\textbf{$\varphi(x_{m}, x_{n})$} & \textbf{BA} &\multicolumn{2}{c}{\textbf{Adversarial Training}} &\textbf{Knowledge Distillation} &\textbf{Descriptions} \\
         & & & \textbf{Clean (\%)} & \textbf{AA (\%)} & \textbf{Top-1 (\%)} & \\
        \midrule
        (a)           &Na        & Na       &62.87 &30.29  &71.03 &baseline with KL loss.   \\  
        \midrule
        (b)           &$\sqrt{\mathcal{S}(\mathbf{w}_{m})}$          &\ding{55} &62.54 &30.20           &71.03          &DKL, equivalent to KL loss.  \\
        (c)           &$\sqrt{\mathcal{S}(\mathbf{w}_{m})}$          &\ding{52} &62.69 &30.42           &71.60         &(b) with BA.  \\
        (d)           &$\sqrt{\mathcal{S}(\mathbf{w}_{m})^{0.4}}$ &\ding{52} &64.15 &31.70           &71.80 &(c) with sample-wise $\varphi^{*}(x_{m}, x_{n})$.  \\
        (e)           &$\sqrt{\mathcal{S}(\mathbf{\bar w}_{m})^{0.4}}$ &\ding{52} &\textbf{65.76} &\textbf{31.91}   &\textbf{71.91} &(c) with class-wise $\varphi^{*}(x_{m}, x_{n})$. \\
        \bottomrule
    \end{tabular} 
    }
    \label{tab:global_info_dkl}
    \vspace{-0.1in}
\end{table*}

\begin{figure*}[tb!]
    \centering
    \subfloat[GKL-AT]            { \includegraphics[width=0.308\linewidth]{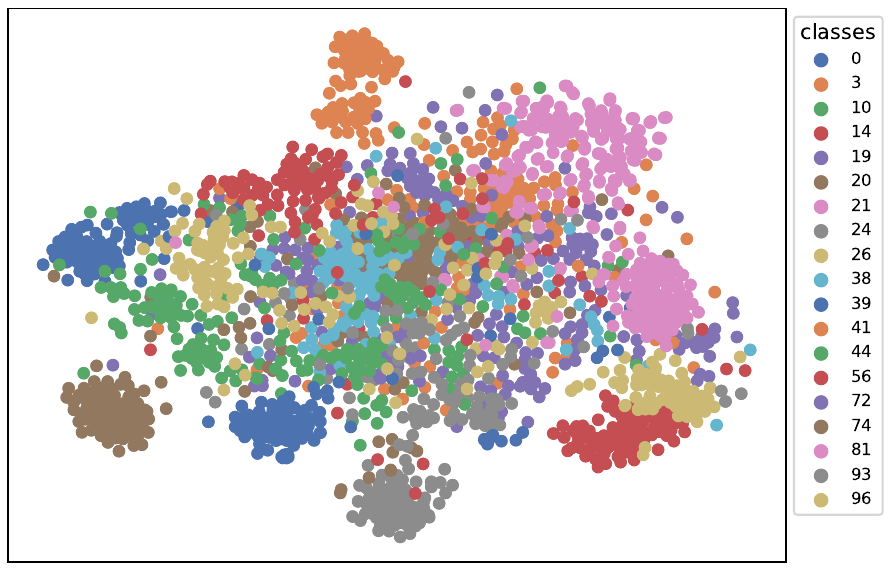} \label{fig:tsne_ikl_at}}
    \subfloat[TRADES]            { \includegraphics[width=0.308\linewidth]{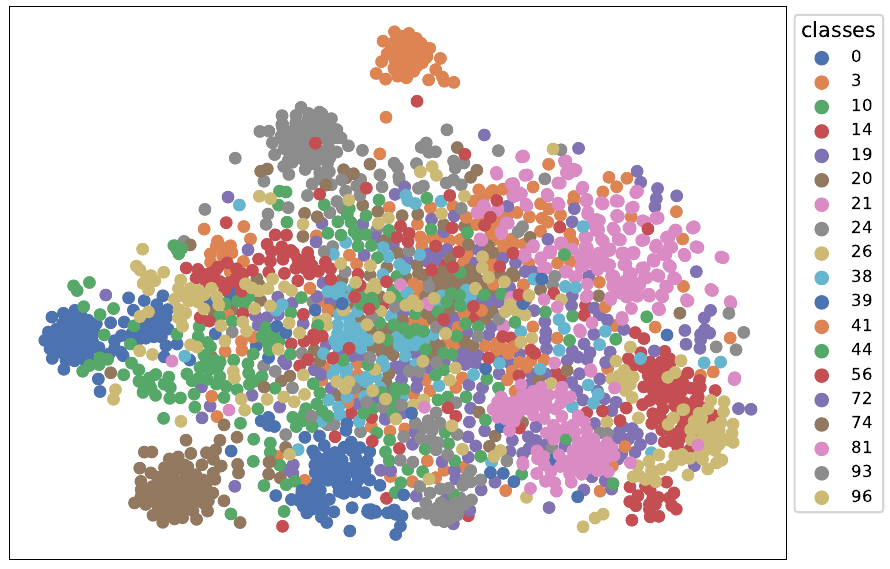}  \label{fig:tsne_trades}}
    \subfloat[Margin differences]{ \includegraphics[width=0.308\linewidth]{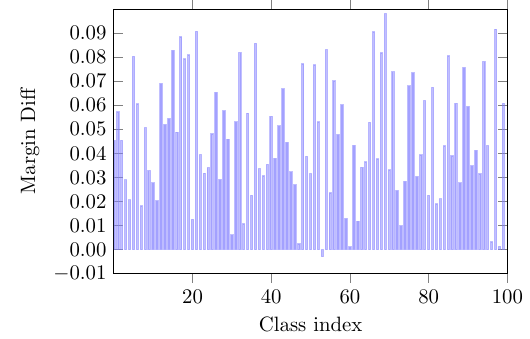}        \label{fig:margin_diff}}
    \caption{
        \textbf{Visualization comparisons.}
        (a) t-SNE visualization of the model trained by GKL-AT on CIFAR-100;
        (b) t-SNE visualization of the model trained by TRADES on CIFAR-100.
        (c) Class margin differences between models trained by GKL-AT and TRADES.
    }
\end{figure*}

\subsection{Generalized Kullback-Leibler Divergence Loss}
\label{sec: IKL}
Based on the analysis in Sec.~\ref{sec:dkl}, we further propose the Generalized Kullback-Leibler (GKL) divergence loss. Compared with DKL in Theorem~\ref{thm:thm_dkl}, GKL introduces two key improvements: (1) breaking the asymmetric optimization property, and (2) designing a more appropriate weighting function $\varphi(x_{m}, x_{n})$. The details are presented below.

\minisection{Breaking the Asymmetric Optimization Property}
As shown in Eq.~\eqref{eq_dkl}, the $\mathbf{w}$MSE component encourages $\mathbf{o}_{n}$ to match $\mathbf{o}_{m}$ by capturing second-order information, namely the pairwise differences between logits of different classes. Each term in $\mathbf{w}$MSE involves only two class logits, and we refer to this property as \textit{locality}. 
In contrast, the Cross-Entropy component in Eq.~\eqref{eq_dkl} encourages $\mathbf{s}_{n}$ to match $\mathbf{s}_{m}$ at the prediction-score level. Each term in the Cross-Entropy depends on all class logits through the softmax distribution, which we refer to as \textit{globality}. 
The two components work together by aligning $\mathbf{o}_{n}$ and $\mathbf{o}_{m}$ from both local and global perspectives. Removing either component can weaken this alignment and lead to performance degradation.

Moreover, we compute the class-wise mean prediction score of models trained on the long-tailed or full ImageNet dataset. As shown in Fig.~\ref{fig:score_entropy}, the prediction scores corresponding to the ground-truth labels are unevenly distributed across classes. 
When applying KL/DKL loss to KD without the $\mathbf{w}$MSE component, classes with higher teacher prediction scores require the student model to produce outputs with much lower entropy than classes with lower teacher prediction scores. This leads to a more difficult optimization problem and can slow convergence, particularly for classes with higher teacher confidence, ultimately degrading performance on those classes.

Unfortunately, due to the asymmetric optimization property of KL loss, an undesirable case can arise when $\mathbf{s}_{m}$ is detached from gradient backpropagation, as in scenarios such as knowledge distillation. In this case, the formulation becomes:
\begin{eqnarray}
    \mathcal{L}_{DKL-KD}(x_m, x_n)\!&\!=\!\underbrace{\frac{\alpha}{4} \|\sqrt{\varphi(x_{m}, x_{n})}(\mathcal{S}(\Delta \mathbf{m}) \!-\! \mathcal{S}(\Delta \mathbf{n}))\|^2}_{\textbf{weighted MSE (wMSE)}} \nonumber\\
    &\underbrace{-\beta \cdot \mathcal{S}(\mathbf{s}_{m}^{\top}) \cdot \log \mathbf{s}_{n}}_{\textbf{Cross-Entropy}}.
    \label{eq:eq_dkl_distill_1}
\end{eqnarray}
As shown in Eq.~\eqref{eq:eq_dkl_distill_1}, the $\mathbf{w}$MSE term no longer contributes to optimization, because all of its sub-components are detached from gradient propagation. Knowledge distillation exactly falls into this case, since the teacher model is fixed throughout training.

Thanks to the decoupled structure of DKL formulation, we address the issue by breaking the asymmetric optimization property, \textit{i.e.}, enabling the gradients of $\mathcal{S}(\Delta \mathbf{n})$ in Eq.~\eqref{eq_dkl},
along with a smoother weight function of $\varphi(x_{m}, x_{n})$.
Then, the updated formulation of Eq.~\eqref{eq:eq_dkl_distill_1} becomes,
\begin{eqnarray}
    \widehat{\mathcal{L}}_{DKL-KD}(x_m, x_n) =& \underbrace{\frac{\alpha}{4} \|\sqrt{\varphi(x_{m}, x_{n})}(\mathcal{S}(\Delta \mathbf{m}) - \Delta \mathbf{n})\|^2}_{\textbf{weighted MSE (wMSE)}} \nonumber\\ 
    &\underbrace{-\beta \cdot \mathcal{S}(\mathbf{s}_{m}^{\top}) \cdot \log \mathbf{s}_{n}}_{\textbf{Cross-Entropy}}.
    \label{eq_dkl_distill_2}
\end{eqnarray}
After enabling the gradients of $\mathcal{S}(\Delta \mathbf{n})$, $\mathbf{w}$MSE will produce symmetric gradients on $o_{n}$ and $o_{m}$. Meanwhile, the smoother $\varphi(x_{m}, x_{n})$ alleviates the problem of hard convergence in classes with higher teacher-predicted scores. It is worth noting that a higher temperature in vanilla KD reduces the risk of hard training convergence and also eliminates the useful ``dark knowledge'', \textit{i.e.}, class relationships. We discuss the designs of $\varphi(x_{m}, x_{n})$ in the following.

\minisection{Proper Design of $\varphi(x_{m}, x_{n})$}
Considering \textit{the hard training convergence problem} and \textit{sample-wise prediction bias}, we propose the sample-wise and class-wise weight function:
\begin{equation}
     \varphi^{*}(x_{m}, x_{n})= \left \{
     \begin{aligned}
      \sqrt{\mathcal{S}(\mathbf{w}_{m})^{\gamma}}, \quad w_{m}^{j,k} = s_{m}^{j} * s_{m}^{k}, \\
      \sqrt{\mathcal{S}(\mathbf{\bar w}_{m})^{\gamma}}, \quad \bar w_{y}^{j,k} = \bar s_{y}^{j} * \bar s_{y}^{k},
     \end{aligned}
     \right. 
\end{equation}
where $\gamma \in [0,1]$ is a smooth factor, $y$ is ground truth label of $x_{m}$, $\mathbf{\bar s}_{y} = \mathbb{E}_{x_{i} \in X_{y}}(s_{i})$, $\mathbf{w}_{m}$ is the sample-wise weight while $\mathbf{\bar w}_{y}$ is the class-wise weight.

As both $0 \leq \mathbf{w}_{m} \leq 1.0$ and $0 \leq \mathbf{\bar w}_{m} \leq 1.0$, $\varphi^{*}(x_{m}, x_{n})$ becomes smoother with $\gamma < 1.0$, facilitating the training convergence of classes with higher teacher-predicted scores. Additionally, the model often cannot output correct predictions when dealing with outliers or hard examples. Then, $\varphi^{*}(x_{m}, x_{n})=\sqrt{\mathcal{S}(\mathbf{w}_{m})^{\gamma}}$ will attach the most importance on the predicted class $\hat{y} = \arg \max o_{m}$ rather than the ground-truth class, which misleads the optimization and makes the training unstable. The class-wise $\varphi^{*}(x_{m},x_{n})$ enhances intra-class consistency and mitigate biases that might arise from sample noises. Especially, in the late stage of training, $\mathbf{\bar w}_{y}$ can always provide correct predictions, benefiting the optimization of $\mathbf{\bar w}$MSE component.    

To this end, we derive the GKL loss in Eq.~\eqref{eq_ikl} by incorporating the two designs,
\begin{eqnarray}
    \mathcal{L}_{GKL}(x_m, x_n) =& \underbrace{\frac{\alpha}{4} \|\sqrt{\varphi^{*}(x_{m},x_{n})}(\Delta \mathbf{m}-\Delta \mathbf{n})\|^2}_{\textbf{weighted MSE ($\mathbf{\bar w}$MSE)}}  \nonumber \\
    &\underbrace{-\beta \cdot \mathcal{S}(\mathbf{s}_{m}^{\top}) \cdot \log \mathbf{s}_{n}}_{\textbf{Cross-Entropy}}.
    \label{eq_ikl}
\end{eqnarray}

\subsection{A Case Study and Analysis}
\label{sec:case_study}

\minisection{A Case Study}
We empirically evaluate each component of GKL on CIFAR-100 for adversarial training and on ImageNet for KD. The ablation results and corresponding experimental settings are summarized in Table~\ref{tab:global_info_dkl}. 
For adversarial training, we adopt the improved TRADES~\cite{zhang2019theoretically} baseline, which incorporates AWP~\cite{wu2020adversarial} and an increasing perturbation budget schedule~\cite{addepalli2022efficient}. For knowledge distillation, we use the official implementation of DKD.
The comparison between (a) and (b) shows that DKL achieves performance comparable to KL, which verifies their theoretical equivalence. Furthermore, the comparisons among (b), (c), (d), and (e) demonstrate the effectiveness of the proposed ``BA'' strategy and the weighting function $\varphi^{*}(x_{m},x_{n})$.

\minisection{Analysis on Class-wise $\varphi^{*}(x_{m}, x_{n})$ for Adversarial Robustness}
As shown in Table~\ref{tab:global_info_dkl}, the class-wise weighting function $\varphi^{*}(x_{m}, x_{n})$ plays an important role in improving adversarial robustness. The class mean probability vector $\bar{\mathbf{s}}_{y}$, computed from all samples in class $y$, is more stable than sample-wise probability vectors. During training, incorrect predictions on hard samples or outliers can cause $\mathbf{w}_{m}$ in Eq.~\eqref{eq_dkl} to provide misleading optimization signals. Replacing it with $\bar{\mathbf{w}}_{y}$ in Eq.~\eqref{eq_ikl} helps alleviate this issue while also enhancing intra-class consistency.

To further illustrate the benefit of incorporating class-wise global information, we define the boundary margin for class $y$ as
\begin{equation}
    \text{Margin}_{y} = \bar{\mathbf{s}}_{y}[y] - \max_{k \neq y}\bar{\mathbf{s}}_{y}[k].
\end{equation}
We then compare the margin differences between models trained by GKL-AT and TRADES on CIFAR-100. As shown in Fig.~\ref{fig:margin_diff}, the margin differences are positive for almost all classes, indicating that GKL-AT learns larger decision boundary margins than TRADES. Such larger margins generally lead to stronger robustness, which is consistent with the results in Sec.~\ref{sec:adv}.

We also randomly sample 20 classes from CIFAR-100 for t-SNE visualization, where the numbers in the figures denote class indices. For each selected class, we collect feature representations of both natural and adversarial examples from the validation set. The visualizations are shown in Figs.~\ref{fig:tsne_trades} and~\ref{fig:tsne_ikl_at}. Compared with TRADES trained using KL loss, features learned by GKL-AT are more compact and better separated.

\minisection{Analysis on Training Convergence for Knowledge Distillation}
A larger temperature can smooth the teacher’s predicted score distribution in vanilla KD, which can help alleviate the training convergence difficulty discussed above. However, using an excessively large temperature can also weaken useful ``dark knowledge,'' \textit{i.e.}, the inter-class relationships that benefit transfer learning. 
As shown in Table~\ref{tab:training_convergence}, the performance on ``Many'' classes with temperatures of 2.0 or 1.5 is noticeably better than that with a temperature of 1.0. This confirms that a smoother teacher distribution can facilitate optimization for classes with high teacher-prediction scores during KD training. 
Meanwhile, GKL-KD consistently achieves substantially better overall performance than KL-KD under different temperature settings, demonstrating the superiority of the proposed weighting function $\varphi^{*}(x_{m}, x_{n})$ in the GKL loss.

\begin{table}[t]
    \centering
    \caption{\textbf{Training convergence analysis on classes with high teacher-predicted scores.} With ResNeXt-101 as the teacher model, the ResNet-18 models are trained on ImageNet-LT.}
    \vspace{-0.1in}
    \resizebox{1.0\linewidth}{!}
    {
    \begin{tabular}{ccccc}
    \toprule
    Method &Many(\%) &Medium(\%) &Few(\%) &All(\%)  \\
    \midrule
    KL-KD~\cite{kd} (temperature=1.0) &64.60 &37.88 &9.53 &44.32 \\
    KL-KD~\cite{kd} (temperature=1.5) &65.39 &37.90 &8.71 &44.51 \\
    KL-KD~\cite{kd} (temperature=2.0) &66.50 &36.70 &6.54 &44.07 \\
    \midrule
    \textbf{GKL-KD} (temperature=1.0) &\textbf{66.72} &\textbf{38.69} &8.69 &\textbf{45.40} \\
    \bottomrule
    \end{tabular}
    }
    \label{tab:training_convergence}
\end{table}

\section{Experiments}
\label{sec:exp}

To verify the effectiveness of our GKL loss, we conduct experiments on CIFAR-10, CIFAR100, ImageNet, and vision-language data for adversarial training (Sec.~\ref{sec:adv}) and knowledge distillation (Sec.~\ref{sec:kd}, Sec.~\ref{sec:imbalanced_kd} and Sec.~\ref{sec:clip}). More ablation studies are included in Sec.~\ref{sec:ablation}.

\subsection{Adversarial Robustness}
\label{sec:adv}

\minisection{Experimental Settings}
We use an improved version of TRADES~\cite{zhang2019theoretically} as our baseline, which incorporates AWP~\cite{wu2020adversarial} and adopts an increasing epsilon schedule. SGD optimizer with a momentum of 0.9 is used. We use the cosine learning rate strategy with an initial learning rate of 0.2 and train models 200 epochs. The batch size is 128, the weight decay is 5e-4 and the perturbation size $\epsilon$ is set to 8/255. Following previous work~\cite{zhang2019theoretically, cui2021learnable}, standard data augmentation including random crops and random horizontal flip is performed for data preprocessing. Models are trained with 4 Nvidia GeForce 3090 GPUs.

Under the setting of training with synthesized data by generative models, we strictly follow the training configurations in DM-AT~\cite{wang2023better} for fair comparisons. Our implementations are based on their open-sourced code. We only replace the KL loss with our GKL loss.

\begin{table*}[tb!]
    \centering
    \caption{
        \textbf{Test accuracy (\%) of clean images and robustness (\%) under AutoAttack on CIFAR-100.}
        All results are the average over three trials.
    }
    \vspace{-0.1in}
    \label{tab:sota_cifar100}
    {
        \begin{tabular}{cclccc}
            \toprule
            Dataset  & Method & Architecture & Augmentation Type & Clean & AA \\
            \midrule
            \multirow{10}{*}{\begin{tabular}{c}\textbf{CIFAR-100}\\
                ($\ell_{\infty}$, $\epsilon=8/255$)
            \end{tabular}}  & AWP~\cite{wu2020adversarial}      & WRN-34-10     & Basic        & 60.38 & 28.86 \\
            & LBGAT~\cite{cui2021learnable}    & WRN-34-10     & Basic        & 60.64 & 29.33 \\
            & LAS-AT~\cite{Jia_2022_CVPR}   & WRN-34-10     & Basic        & 64.89  & 30.77 \\
            & ACAT~\cite{addepalli2022efficient}     & WRN-34-10     & Basic        & 65.75  & 30.23 \\
            & \textbf{GKL-AT}          & WRN-34-10    & Basic        & \textbf{65.76} & \textbf{31.91} \\
            \cmidrule{2-6}
            & ACAT~\cite{addepalli2022efficient}      & WRN-34-10     & AutoAug        & \textbf{68.74}  & 31.30 \\
            & \textbf{GKL-AT}                         & WRN-34-10     & AutoAug        &66.08  & \textbf{32.53} \\
            \cmidrule{2-6}
            &DM-AT~\cite{wang2023better} & WRN-28-10     & 50M Generated Data    & 72.58 & 38.83 \\
            & \textbf{GKL-AT}       & WRN-28-10     & 50M Generated Data    & \textbf{73.65} &\textbf{39.37} \\
            \bottomrule
        \end{tabular}%
    }
\end{table*}%

\begin{table*}[t]
\centering
\caption{
    \textbf{Test accuracy (\%) of clean images and robustness (\%) under AutoAttack on CIFAR-10.} Average over three trials are listed.
}
\vspace{-0.1in}
{
\renewcommand*{\arraystretch}{1.0}
\begin{tabular}{cclccc}
\toprule
Dataset  & Method & Architecture & Augmentation Type & Clean & AA \\
\midrule
\multirow{10}{*}{\begin{tabular}{c}\textbf{CIFAR-10} \\
    ($\ell_{\infty}$, $\epsilon=8/255$)
                \end{tabular}}        & Rice et al.~\cite{rice2020overfitting}           & WRN-34-20       & Basic  & 85.34 & 53.42 \\
                                      & LBGAT~\cite{cui2021learnable}        & WRN-34-20       & Basic  & \textbf{88.70} & 53.57 \\
                                      & AWP~\cite{wu2020adversarial}          & WRN-34-10       & Basic  & 85.36 & 56.17 \\ 
                                      & LAS-AT~\cite{Jia_2022_CVPR}       & WRN-34-10       & Basic  & 87.74 & 55.52 \\
                                      & ACAT~\cite{addepalli2022efficient}         & WRN-34-10       & Basic  & 82.41 & 55.36 \\
                                      & \textbf{GKL-AT}                      & WRN-34-10       & Basic  & 84.80 & \textbf{57.09} \\
                                      \cmidrule{2-6}
                                      & ACAT~\cite{addepalli2022efficient}    & WRN-34-10       & AutoAug  & 88.64 & 57.05 \\
                                      & \textbf{GKL-AT}                      & WRN-34-10       & AutoAug  & 85.20 & \textbf{57.62} \\
                                      \cmidrule{2-6}
                                   
                                      & DM-AT~\cite{wang2023better}                & WRN-28-10       & 20M Generated Data    & 92.44 & 67.31 \\ 
                                      & \multirow{1}{*}{\textbf{GKL-AT}}     & WRN-28-10       & 20M Generated Data   &92.16 &\textbf{67.75} \\
\bottomrule
\end{tabular}%
\label{tab:sota_cifar10}
}
\end{table*}%

\minisection{Datasets and Evaluation}
Following previous work~\cite{wu2020adversarial, cui2021learnable}, CIFAR-10 and CIFAR-100 are used for the adversarial training task. we report the clean accuracy on natural images and adversarial robustness under Auto-Attack~\cite{croce2020reliable} with epsilon 8/255. 

\begin{table*}[tb!]
    \center
    \caption{\textbf{Top-1 accuracy~(\%) on the ImageNet validation and training speed (sec/iteration) comparisons.} Training speed is calculated on 4 Nvidia GeForce 3090 GPUs with a batch of 512 224x224 images.
    All results are the average over three trials.
    }
    \vspace{-0.1in}
{
\begin{tabular}{ccccccc}
\toprule
\multirow{4}{*}{\begin{tabular}[c]{@{}c@{}}Distillation \\ Manner\end{tabular}} & \multirow{2}{*}{Teacher}  & \multirow{4}{*}{Extra Parameters} & \multicolumn{2}{c}{ResNet34}  & \multicolumn{2}{c}{ResNet50} \\
&      & & \multicolumn{2}{c}{73.31} &\multicolumn{2}{c}{76.16}   \\
&  \multirow{2}{*}{Student}  & & \multicolumn{2}{c}{ResNet18} & \multicolumn{2}{c}{MobileNet} \\
& \space     & & \multicolumn{2}{c}{69.75} & \multicolumn{2}{c}{68.87} \\
\midrule
\multirow{4}{*}{Features}
& AT~\cite{at}             & \ding{55} & 70.69   &                 & 69.56    &\\
& OFD~\cite{ofd}            & \ding{52} & 70.81   &                 & 71.25    &\\
& CRD~\cite{crd}            & \ding{52} & 71.17   &                 & 71.37    &\\ 
& ReviewKD~\cite{revkd}       & \ding{52} & 71.61   &0.319 s/iter     & 72.56    &0.526 s/iter\\ 
\midrule
\multirow{4}{*}{Logits}   
& DKD~\cite{dkd}         & \ding{55} & 71.70                            &                 & 72.05 &\\
& KD~\cite{kd}          & \ding{55} & 71.03                            &                 & 70.50 &\\
& IKL-KD~\cite{cui2025decoupled}        & \ding{55}  &\textbf{71.91} &\textbf{0.197 s/iter} &72.84 &\textbf{0.252 s/iter} \\
& \textbf{GKL-KD}      & \ding{55} & \textbf{71.91}                   &\textbf{0.197 s/iter}     & \textbf{72.92}                 &\textbf{0.252 s/iter}  \\
\bottomrule
\end{tabular}
}
\label{tab:imagenet_kd}
\end{table*}

\begin{table*}[tb!]
\setlength{\tabcolsep}{4.0mm}
\center
\caption{\textbf{Peformance~(\%) on imbalanced data, \textit{i.e.}, the ImageNet-LT.}}
\vspace{-0.1in}
{
\begin{tabular}{ccccccc}
\toprule
Method   & Teacher &Student  &Many(\%) &Medium(\%) &Few(\%) &All(\%)\\
\midrule
Baseline & - &ResNet-18  &63.16 &33.47 &5.88 &41.15 \\
Baseline & - &ResNet-50  &67.25 &38.56 &8.21 &45.47 \\
Baseline & - &ResNet-101 &68.91 &42.32 &11.24 &48.33 \\
\midrule
KL-KD~\cite{kd}    &ResNeXt-101  &ResNet-18 &64.6 &37.88 &9.53  &44.32 \\
KL-KD~\cite{kd}    &ResNeXt-101  &ResNet-50 &68.83 &42.31 &11.37 &48.31 \\
\midrule
IKL-KD~\cite{cui2025decoupled}             &ResNeXt-101  &ResNet-18 &66.60 &38.53 &8.19 &45.21 \\
IKL-KD~\cite{cui2025decoupled}             &ResNeXt-101  &ResNet-50 &70.06 &43.47 &10.99 &49.29 \\
\textbf{GKL-KD}    &ResNeXt-101  &ResNet-18 &66.72 &38.69 &8.69  &\textbf{45.40} \\
\textbf{GKL-KD}    &ResNeXt-101  &ResNet-50 &70.31 &43.47 &10.85 &\textbf{49.40} \\
\bottomrule
\end{tabular}
}
\label{tab:imagenetlt_kd}
\end{table*}

\begin{table}[htp]
    \centering
    \caption{\textbf{Zero-shot Top-1 performance (\%) of ClIP models with knowledge distillation.} Various temperatures are applied to traditional knowledge distillation with KL loss.}
    \vspace{-0.1in}
    \resizebox{1.0\linewidth}{!}
    {
    \begin{tabular}{ccccc}
         \toprule
         Method &  ImageNet & ImageNet-V2 & ImageNet-Sketch & ImageNet-R \\
         \midrule
         \multicolumn{5}{c}{Teacher: pre-trained OpenCLIP ViT-L/14} \\
         \midrule
         - & 79.2 & - &- &-\\ 
         \midrule
          \multicolumn{5}{c}{Student: ViT-B/16 training from scratch} \\
         \midrule
         Baseline   &53.21 &45.21 &37.89 &55.95\\
         KL-KD~\cite{kd}(t=1) &57.28 &49.18 &42.15 &62.24 \\
         KL-KD~\cite{kd}(t=2) &57.71 &49.73 &43.24 &63.26 \\
         KL-KD~\cite{kd}(t=4) &59.17 &50.97 &45.48 &66.86 \\
         KL-KD~\cite{kd}(t=8) &60.85  &53.20 &45.87 &67.50\\
         \midrule
         \textbf{GKL-KD}(t=8)   &\textbf{61.62}  &\textbf{53.73} &\textbf{46.64} &\textbf{68.70}\\
         \bottomrule
    \end{tabular}
    }
    \label{tab:clip_dist}
\end{table}

\begin{table*}[tb!]
    \centering
    \caption{\textbf{LLaVA performance with GKL-KD models.}}
    \vspace{-0.1in}
    \begin{tabular}{ccccccccc}
    \toprule
     \multirow{2}{*}{CLIP Model}  &\multirow{2}{*}{TextVQA} &\multirow{2}{*}{MMBench} &\multirow{2}{*}{MMBench-CN} &\multicolumn{2}{c}{GQA} &\multicolumn{3}{c}{POPE}  \\
     \cmidrule{5-9}
     & & & &Open &Accuracy &Adversarial &Popular & Random \\
     \midrule
     Baseline    &45.83          &49.66 &39.86 &55.29 &39.71 &75.05 &81.43 &51.75 \\
     KL-KD~\cite{kd}       &\textbf{47.63} &51.89 &42.01 &57.28 &41.72 &\textbf{77.87} &82.33 &50.76 \\
     \midrule
     \textbf{GKL-KD}      &47.51 &\textbf{52.06} &\textbf{43.38} &\textbf{57.66} &\textbf{42.29} &77.80 &\textbf{83.07} &\textbf{51.86} \\
     \bottomrule
    \end{tabular}
    \label{tab:llava}
    \vspace{-0.1in}
\end{table*}

\begin{table*}[tb!]
\centering
\caption{\textbf{Ablation study on hyper-parameters of GKL.}}
\vspace{-0.1in}
\begin{minipage}{0.24\textwidth}
\centering
\renewcommand{\arraystretch}{1.2}
{
\setlength{\tabcolsep}{3.5mm}
\begin{tabular}{ccc}
\toprule
$\frac{\alpha}{4}$ & Clean  & AA\\
\midrule
    $\frac{\alpha}{4}=3$     & 67.52 &31.29 \\
    $\frac{\alpha}{4}=4$     & 66.26 &31.33 \\ 
    $\frac{\alpha}{4}=5$     & 65.76 &31.91 \\ 
    $\frac{\alpha}{4}=6$     & 65.14 &31.64 \\ 
\bottomrule
\end{tabular}
}
\caption{Effects of $\frac{\alpha}{4}$.}
\label{tab:ablation_alpha}
\end{minipage}
\hspace{0.01in}
\begin{minipage}{0.24\textwidth}
\centering
\vspace{-0.01in}
\renewcommand{\arraystretch}{1.2}
{
\setlength{\tabcolsep}{3.5mm}
\begin{tabular}{ccc}
\toprule
\textbf{$\beta$} & Clean & AA \\
\midrule
    $\beta=2$     & 66.13 &30.95 \\
    $\beta=3$     & 66.31 &31.33 \\
    $\beta=4$     & 66.00 &31.57 \\  
    $\beta=5$     & 65.76 &31.91 \\ 
\bottomrule
\end{tabular}
}
\caption{Effects of $\beta$.}
\label{tab:ablation_beta}
\end{minipage}
\hspace{0.015in}
\begin{minipage}{0.32\textwidth}  
\centering
\vspace{+0.01in}
\renewcommand{\arraystretch}{1.1}
{
\setlength{\tabcolsep}{3.0mm}
\begin{tabular}{ccc}
\toprule
\textbf{$\gamma$} & Clean & AA \\
\midrule
    $\gamma=0.4$ (sample-wise)   &64.15 &31.70 \\
    \midrule
    $\gamma=0.2$ (class-wise)    & 66.69 &31.23 \\
    $\gamma=0.3$ (class-wise)    & 65.80 &31.59 \\
    $\gamma=0.4$ (class-wise)    & 65.76 &31.91 \\  
\bottomrule
\end{tabular}
}
\caption{Effects of $\gamma$.}
\label{tab:ablation_gamma_adv}
\end{minipage}
\label{tab:ablation_alpha_beta_tau}
\vspace{-0.2in}
\end{table*}

\minisection{Comparison Methods}
To compare with previous methods, we categorize them into two groups according to the different types of data preprocessing:
\begin{itemize}
    \item Methods with basic augmentation, \textit{i.e.}, random crops and random horizontal flip.
    \item Methods using augmentation with generative models or Auto-Aug~\cite{DBLP:conf/cvpr/CubukZMVL19}, CutMix~\cite{yun2019cutmix}. 
\end{itemize}

\minisection{Comparisons with State-of-the-art on CIFAR-100}
On CIFAR-100, with the basic augmentations setting, we compare with AWP, LBGAT, LAS-AT, and ACAT.
The experimental results are summarized in Table~\ref{tab:sota_cifar100}. Our WRN-34-10 models trained with GKL loss do a better trade-off between natural accuracy and adversarial robustness. With basic augmentation, our model achieves \textbf{65.76\%} top-1 accuracy on natural images while \textbf{31.91\%} adversarial robustness under Auto-Attack. 
An interesting phenomenon is that GKL-AT is complementary to data augmentation strategies, like AutoAug, without any specific designs, which is different from the previous observation that the data augmentation strategy hardly benefits adversarial training~\cite{wu2020adversarial}. 
With AutoAug, we obtain \textbf{32.53\%} adversarial robustness, achieving a new state-of-the-art under the setting without extra real or generated data.  

We follow DM-AT~\cite{wang2023better} to take advantage of synthesized images generated by the popular diffusion models~\cite{karras2022elucidating}.  
With 50M generated images, we create a new state-of-the-art with WideResNet-28-10, achieving \textbf{73.65\%} top-1 natural accuracy and \textbf{39.37\%} adversarial robustness under Auto-Attack.

\minisection{Comparison with State-of-the-art on CIFAR-10}
Experimental results on CIFAR-10 are listed in Table~\ref{tab:sota_cifar10}. With the basic augmentation setting,
our model achieves 84.80\% top-1 accuracy on natural images and 57.09\% robustness,
outperforming AWP by 0.92\% on robustness.
With extra generated data, we improve the state-of-the-art by 0.44\%, achieving \textbf{67.75\%} robustness.

\subsection{Knowledge Distillation on Balanced Data}
\label{sec:kd}

\minisection{Datasets and Evaluation}
Following previous work~\cite{revkd,crd}, we conduct experiments on CIFAR-100~\cite{cifar} and ImageNet~\cite{imagenet} to show the advantages of GKL on knowledge distillation.
For evaluation, we report top-1 accuracy on CIFAR-100 and ImageNet validation. The training speed of different methods is also discussed.

\minisection{Experimental Settings}
We follow the experimental settings in DKD. Our implementation for KD is based on their open-sourced code. Models are trained with 1 and 8 Nvidia GeForce 3090 GPUs on CIFAR and ImageNet separately. 

Specifically, on CIFAR-100, we train models for 240 epochs with a learning rate that decayed by 0.1 at the 150th, 180th, and 210th epoch. We initialize the learning rate to 0.01 for MobileNet and ShuffleNet, and 0.05 for other models. The batch size is 64. We train all models three times and report the mean accuracy. 
On ImageNet, we use the standard training that trains the model for 100 epochs and decays the learning rate for every 30 epochs. We initialize the learning rate to 0.2 and set the batch size to 512.

For both CIFAR-100 and ImageNet, we consider the distillation among the architectures having the same unit structures, like ResNet56 and ResNet20, VGGNet13 and VGGNet8. On the other hand, we also explore the distillation among architectures made up of different unit structures, like WideResNet and ShuffleNet, VggNet.

\minisection{Comparison Methods}
According to the information extracted from the teacher model in distillation training, KD methods can be divided into two categories: 
\begin{itemize} 
    \item Features-based methods~\cite{fitnets,crd, revkd, ofd}. This kind of method makes use of features from different layers of the teacher model, which can need extra parameters and high training computational costs.
    \item Logits-based methods~\cite{kd, dkd}. This kind of method only makes use of the logits output of the teacher model, which does not require knowing the architectures of the teacher model and thus is more general in practice.
\end{itemize}

\minisection{Comparison with State-of-the-art on CIFAR-100}
Experimental results on CIFAR-100 are summarized in Table~\ref{tab:cifar_kd} and Table~\ref{tab:cifar2_kd} (in Appendix).
Table~\ref{tab:cifar_kd} lists the comparisons with previous methods under the setting that the architectures of the teacher and student have the same unit structures. Models trained by GKL-KD can achieve comparable or better performance in all considered settings. Specifically, we achieve the best performance in 4 out of 6 training settings.
Table~\ref{tab:cifar2_kd} shows the comparisons with previous methods under the setting that the architectures of the teacher and student have different unit structures. We achieve the best performance in 3 out of 5 training configurations.

\minisection{Comparison with State-of-the-art on ImageNet}
We empirically show the comparisons with other methods on ImageNet in Table~\ref{tab:imagenet_kd}.
With a ResNet34 teacher, our ResNet18 achieves \textbf{71.91\%} top-1 accuracy. With a ResNet50 teacher, our MobileNet achieves \textbf{72.92\%} top-1 accuracy. Models trained by GKL-KD surpass all previous methods while saving \textbf{38\%} and \textbf{52\%} computation costs for ResNet34--ResNet18 and ResNet50--MobileNet distillation training respectively when compared with ReviewKD~\cite{revkd}.

\subsection{Knowledge Distillation on Imbalanced Data}
\label{sec:imbalanced_kd}
Data often follows a long-tailed distribution. Tackling the long-tailed recognition problem is essential for real-world applications. Lots of research has contributed to algorithms and theories~\cite{cao2019learning, cui2019class, kang2019decoupling, menon2020long,cui2022generalized,cui2021parametric, cui2024classes} on the problem.
In this work, we examine how the knowledge distillation with our GKL loss affects model performance on imbalanced data, \textit{i.e.}, ImageNet-LT~\cite{liu2019large}. We train ResNets models 90 epochs with \textit{Random-Resized-Crop} and horizontal flip as image pre-processing. Following previous work~\cite{9774921}, we report the top-1 accuracy on Many-shot, Medidium-shot, Few-shot, and All classes.

\noindent{\bf Results Analysis.} As shown in Table~\ref{tab:imagenetlt_kd}, GKL-KD consistently outperforms KL-KD on imbalanced data. Specifically, GKL-KD achieves improvements of 1.48\%, 1.16\% on Many-shot and Medium-shot, respectively, over KL-KD with the teacher-student configuration of ResNeXt-101-ResNet50. This is coherent to our analysis in Section~\ref{sec:case_study}: high-frequency class prediction scores from the teacher model are often higher than those of low-frequency classes, resulting in hard training convergence for high-frequency classes in vanilla KD. GKL-KD addresses this problem with the smooth weighted $\varphi^{*}(\cdot)$. Note that it is contrary to previous work that often focuses on improving low-frequency class performance.

\subsection{Knowledge Distillation on CLIP Models}
\label{sec:clip}

\minisection{\bf CLIP Models} 
To demonstrate the generalizability of our GKL loss, we conduct experiments on vision-language data for CLIP knowledge distillation. CLIP models are trained with image-text pairs using contrastive learning. As there is no parameterized linear classifier, we use the similarity between image and text representations as logits and adopt the sample-wise weight function $\varphi^{*}(x_{m}, x_{n})= \sqrt{\mathcal{S}(\mathbf{w}_{m})^{\gamma}}$ during distillation training. Specifically, we use ViT-B/16 as the student while the pretrained OpenCLIP model ViT-L/14 as the teacher. We train models 32 epochs with a total batch size of 8192 on 15M data which is randomly sampled from DataComp1B for each epoch. Open-sourced code from OpenCLIP~\cite{cherti2023reproducible} is used. The experimental results are summarized in Table~\ref{tab:clip_dist}.

\minisection{Auto-regressive Vision-language Models}
CLIP serves as a fundamental component of multi-modal large language models (MLLMs). Using CLIP as the vision encoder in LLaVA~\cite{liu2023llava}, we investigate the impact of GKL-KD models on MLLM performance. Our study leverages the open-source LLaVA~\cite{liu2023llava}, replacing only the CLIP vision encoder with our models. Specifically, the Vicuna-7B with LoRA is adopted for the LLM backbone. To evaluate the trained models, we employ multiple widely used benchmarks, including TextVQA, POPE, GQA, MMBench, and MMBench-CN. The experimental results are listed in Table~\ref{tab:llava}. In particular, the baseline uses the CLIP model trained on 15M data without KD, while KL-KD and GKL-KD models are trained taking OpenCLIP ViT-L/14 as the teacher.

\subsection{Ablation Studies}
\label{sec:ablation}

\minisection{Ablation on $\gamma$ for Knowledge Distillation}
As $\gamma \in [0,1]$, the weights by function $\varphi^{*}(x_{m}, x_{n})$ becomes smoother, mitigating the training convergence difficulties. As shown in Table~\ref{tab:ablation_gamma}, we conduct experiments on ImageNet-LT with ResNet-18 as the student and ResNeXt-101 as the teacher. Sample-wise $\varphi(x_{m}, x_{n})$ and Class-wise $\varphi^{*}(x_{m}, x_{n})$ weight functions both can improve the performance of the student model. Especially, with class-wise global information, $\varphi^{*}(x_{m}, x_{n})$ further enhances model generalization ability and robustness.

\minisection{Ablation on $\gamma$ for Adversarial Robustness}
To examine the influence of $\gamma$ on adversarial training, we conduct experiments on CIFAR-100 using different $\gamma$ values. The results are summarized in Table~\ref{tab:ablation_gamma_adv}. An interesting phenomenon is observed: First, a smoother \textit{class-wise} weighting function $\varphi(\cdot)$ generally leads to higher clean accuracy, as reflected by the performance gain compared to \textit{sample-wise} weighting, confirming the rational of incorporating global information. Second, as $\gamma$ increases for \textit{class-wise} weighting, adversarial robustness improves while the clean accuracy degrades. This suggests that moderate values of $\gamma$ provide a better balance between clean accuracy and robustness, whereas overly large $\gamma$ may harm robustness due to the accompanying drop in clean accuracy. Please refer to Appendix~6.3 for more empirical analysis.

\begin{table}[tb!]
    \centering
    \caption{\textbf{Ablation of hyper-parameter $\gamma$ on ImageNet-LT.}}
    \vspace{-0.1in}
    \begin{tabular}{cccc}
         \toprule
         $\gamma$ &ImageNet-LT &ImageNet-R &ImageNet-sketch \\
         \midrule
         -(KL-KD) &44.32   &20.94 &9.16 \\
         \midrule
         \multicolumn{4}{c}{Sample-wise $\varphi(x_{m},x_{n})$} \\
         \midrule
         $\gamma=0.0$ &45.19  &21.49 &9.70\\
         $\gamma=0.3$ &44.76  &20.22 &8.94 \\
         \midrule
         \multicolumn{4}{c}{Class-wise $\varphi(x_{m},x_{n})$} \\
         \midrule
         $\gamma=1.0$ &44.62  &20.65 &9.16 \\
         $\gamma=0.5$ &45.28  &21.29 &9.55 \\
         $\gamma=0.3$ &\textbf{45.40}  &\textbf{21.58} &\textbf{9.70} \\
         \bottomrule
    \end{tabular}
    \label{tab:ablation_gamma}
    \vspace{-0.1in}
\end{table}

\begin{table*}[tb!]
\centering
\caption{\textbf{Ablation study of $\epsilon$.}}
\vspace{-0.1in}
{
\setlength{\tabcolsep}{4.1mm}
\begin{tabular}{ccccccccc}
\toprule
\multirow{2}{*}{Method} &\multirow{2}{*}{Clean}  &\multicolumn{7}{c}{AA} \\
\cmidrule{3-9} 
& &$\frac{2}{255}$ &$\frac{4}{255}$ & $\frac{6}{255}$ & $\frac{8}{255}$ & $\frac{10}{255}$ &$\frac{12}{255}$ &Avg.\\
\midrule
   TRADES &62.87 &53.88 &45.31 &37.28 &30.29 &24.28 &19.17 &35.04\\
   \textbf{GKL-AT} &\textbf{63.40} &\textbf{55.31} &\textbf{46.76} &\textbf{38.98} &\textbf{31.91} &\textbf{25.33} &\textbf{19.98} &\textbf{36.38}\\
\bottomrule
\end{tabular}
}
\label{tab:ablation_epsilon}
\end{table*}

\begin{table*}[tb!]
\centering
\caption{\textbf{Evaluation under PGD and CW attacks.}}
\vspace{-0.1in}
{
\begin{tabular}{cccccccc}
\toprule
 Method &Acc &PGD-10 &PGD-20 &CW-10 &CW-20 &Auto-Attack &Worst \\
\midrule
KL-AT(TRADES)  &62.87 &36.01 &35.84 &40.03 &39.86 &30.29 &30.29\\
\midrule
\textbf{GKL-AT(Ours)}                      &63.40 &36.78  &36.55 &40.72 &40.47 &31.91 &31.92\\
\textbf{GKL-AT (Ours with autoaug)}        &65.93 &38.15  &37.75 &41.10 &40.86 &32.53 &32.52\\
\textbf{GKL-AT (Ours with synthetic data)} &73.85 &44.43  &44.12 &47.59 &47.53 &39.18 &39.18 \\
\bottomrule
\end{tabular}
}
\label{tab:pgd_cw}
\end{table*}

\minisection{Ablation on $\alpha$ and $\beta$ for Adversarial Robustness.}
Thanks to the decoupled structure of the DKL loss formulation, the two components of GKL (\textit{i.e.}, $\mathbf{w}$MSE and Cross-Entropy) can be adjusted independently. We empirically study the effects of the hyper-parameters $\alpha$ and $\beta$ on CIFAR-100 for adversarial robustness. Clean accuracy on natural data and robustness under AA~\cite{croce2020reliable} are reported in Table~\ref{tab:ablation_alpha} and Table~\ref{tab:ablation_beta}, respectively. In practice, appropriate choices of $\alpha$ and $\beta$ provide a favorable trade-off between natural accuracy and adversarial robustness. To determine $\beta$, we follow the standard weighting of the KL loss adopted in prior methods. For example, the KL regularization coefficient is commonly set to $5.0$ in ACAT and TRADES; accordingly, we use $\beta = 5.0$ as a default choice. For $\alpha/4$, we perform a grid search over the range $[1,4]$ to select the optimal value.

\minisection{Ablation on Various Perturbation Size $\epsilon$}
We evaluate model robustness with unknown perturbation size $\epsilon$ in training under Auto-Attack.
The experimental results are summarized in Table~\ref{tab:ablation_epsilon}.
As shown in Table~\ref{tab:ablation_epsilon}, model robustness decreases significantly as the $\epsilon$ increases for both the TRADES model and our model. Nevertheless, our model achieves stronger robustness than the TRADES model under all of $\epsilon$, outperforming TRADES by 1.34\% on average robustness, demonstrating the super advantages of models adversarially trained with our GKL loss.

\minisection{Robustness under Other Attacks}
Auto-Attack is currently one of the strongest attack methods. It ensembles several adversarial attack methods including APGD-CE, APGD-DLR, FAB, and Square Attack. To show the effectiveness of our GKL loss, we also evaluate our models under PGD and CW attacks with 10 and 20 iterations. The perturbation size and step size are set to 8/255 and 2/255 respectively. As shown in Table~\ref{tab:pgd_cw}, with increasing iterations from 10 to 20, our models show similar robustness, demonstrating that our models don't suffer from obfuscated gradients problem. 

\minisection{Connection to LBGAT~\cite{cui2021learnable}}
LBGAT~\cite{cui2021learnable} guides the optimization of adversarial training with an extra classification boundary from a naturally trained model. It achieves stronger adversarial robustness meanwhile much better performance on natural images, implying the significance of assistance from a good classification boundary. However, LBGAT requires that the target robust model and the naturally trained model should be optimized simultaneously. It takes additional computation costs and memory consumption.
GKL-AT advances LBGAT in the following aspects. 
\begin{itemize}
    \item With the introduced global information in Sec.~\ref{sec: IKL}, GKL-AT uses the class-wise classification boundary to guide the training optimization, which is different from LBGAT that uses the sample-wise classification boundary from an extra naturally trained model.  
    \item GKL loss as an improved version of KL loss is used for boundary guidance constraints while MSE loss is applied in the LBGAT method.
\end{itemize}

\section{Conclusion and Limitation}
\label{sec:conclusion}
In this paper, we investigated the optimization mechanism of the Kullback-Leibler (KL) divergence loss from a gradient perspective. We showed that KL can be decoupled into a weighted Mean Squared Error ($\mathbf{w}$MSE) term and a Cross-Entropy term with soft labels, leading to the proposed Decoupled Kullback-Leibler (DKL) divergence loss. 
To address the limitations of KL/DKL, we further introduced two improvements: breaking the asymmetric optimization property and designing smoother weighting functions with class-wise global information, resulting in the proposed Generalized Kullback-Leibler (GKL) divergence loss. 
Extensive experiments on CIFAR-10/100, ImageNet, and vision-language benchmarks show that GKL achieves state-of-the-art adversarial robustness and competitive knowledge distillation performance. 
Since the proposed framework only relies on softmax outputs, it can naturally extend to token-level distillation in autoregressive sequence models. Exploring such language-model distillation settings, as well as applications in out-of-distribution robustness and incremental learning, will be valuable future work.

	\ifCLASSOPTIONcaptionsoff
	\newpage
	\fi

	{\small
		\bibliographystyle{IEEEtran}
		\bibliography{egbib}
	}

\onecolumn
\section{Appendix}

\subsection{New state-of-the-art robustness on CIFAR-100/10}
\vspace{+0.1in}
\textit{Robustbench} is the most popular benchmark for adversarial robust models in the community. It evaluates the performance of models by the Auto-Attack. Auto-Attack \cite{croce2020reliable} is an ensemble of different kinds of attack methods and is considered the most effective method to test the robustness of models.

We achieve new state-of-the-art robustness on CIFAR-10 and CIFAR-100 under both settings w/ and w/o generated data.
As shown in Table~\ref{tab:comparision_sota}, on CIFAR-100 without extra generated data, we achieve 32.53\% robustness, outperforming the previous best result by \textbf{0.68\%} while saving \textbf{33.3\%} computational cost. With generated data, our model boosts performance to 73.65\% natural accuracy, surpassing the previous best result by \textbf{1.07\%} while maintaining the \textbf{strongest robustness}. More detailed comparisons can be accessed on the public leaderboard \url{https://robustbench.github.io/}.

\begin{table*}[tb!]
\centering
\caption{\textbf{New state-of-the-art on public leaderboard { \textit RobustBench} \cite{croce2020reliable}.}}
\label{tab:comparision_sota}
{
\begin{tabular}{ccccc}
\toprule
 Experimental Settings & augmentation strategy & Clean  & AA &Computation saving\\
\midrule
    w/o Generated Data (Previous best results)        & Basic   &62.99  &31.20 &\\
    w/o Generated Data (Ours)                         & Basic   &\textbf{65.76(+2.67)}  &\textbf{31.91(+0.71)} &\textbf{33.3\%} \\
    \midrule
    w/o Generated Data (Previous best results)        & Autoaug   &\textbf{68.75}  &31.85 &\\
    w/o Generated Data (Ours)                         & Autoaug   &66.08  &\textbf{32.53(+0.68)} &\textbf{33.3\%}\\
    \midrule
    w/ Generated Data (Previous best results)    & Genreated data &72.58  &38.83 &\\ 
    w/ Generated Data (Ours)                     & Generated data &\textbf{73.65(+1.07)}  &\textbf{39.37(+0.54)} &0\%\\ 
\bottomrule
\end{tabular}
}
\end{table*}

\begin{table*}[tb!]
\center
\caption{\textbf{Top-1 accuracy~(\%) on the CIFAR-100 validation.} Teachers and students are in the \textbf{same} architectures. All results are the average over three trials.}
{
\begin{tabular}{cccccccc}
\toprule
\multirow{4}{*}{\begin{tabular}[c]{@{}c@{}}Distillation \\ Manner\end{tabular}} & \multirow{2}{*}{Teacher}  & ResNet56 & ResNet110 & ResNet32$\times$4 & WRN-40-2 & WRN-40-2 & VGG13 \\
&      & 72.34      & 74.31       & 79.42        & 75.61      & 75.61      & 74.64   \\
& \multirow{2}{*}{Student}  & ResNet20 & ResNet32 & ResNet8$\times$4  & WRN-16-2 & WRN-40-1 & VGG8  \\
& \space     & 69.06      & 71.14       & 72.50        & 73.26      & 71.98      & 70.36   \\
\midrule
\multirow{5}{*}{Features}
& FitNet~\cite{fitnets}   & 69.21      & 71.06       & 73.50        & 73.58      & 72.24      & 71.02   \\
& RKD          & 69.61      & 71.82       & 71.90        & 73.35      & 72.22      & 71.48   \\
& CRD~\cite{crd}          & 71.16      & 73.48       & 75.51        & 75.48      & 74.14      & 73.94   \\
& OFD~\cite{ofd}          & 70.98      & 73.23       & 74.95        & 75.24      & 74.33      & 73.95   \\
& ReviewKD~\cite{revkd}   & 71.89      & 73.89       & 75.63        & 76.12      & \textbf{75.09}      & 74.84   \\ 
\midrule
\multirow{4}{*}{Logits}                                                         
& DKD~\cite{dkd}        & \textbf{71.97}      & 74.11       & 76.32              & 76.24          & 74.81      & 74.68 \\
& KD~\cite{kd}          & 70.66               & 73.08                & 73.33              & 74.92          & 73.54      & 72.98   \\
& IKL-KD~\cite{cui2025decoupled}         &71.44 &\textbf{74.26} &76.59 &\textbf{76.45} &\textbf{74.98} &\textbf{74.98} \\
& \textbf{GKL-KD}      & 71.67               & \textbf{74.26}                & \textbf{76.83}     & \textbf{76.45} & 74.98      & \textbf{74.98} \\
\bottomrule
\end{tabular}
}
\label{tab:cifar_kd}
\end{table*}

\begin{table*}[tb!]
\center
\caption{\textbf{Top-1 accuracy~(\%) on the CIFAR-100 validation.} Teachers and students are in \textbf{different} architectures. All results are the average over 3 trials.}
{
\begin{tabular}{ccccccc}
\toprule
\multirow{4}{*}{\begin{tabular}[c]{@{}c@{}}Distillation \\ Manner\end{tabular}} & \multirow{2}{*}{Teacher}  & ResNet32$\times$4 & WRN-40-2 & VGG13 & ResNet50 & ResNet32$\times$4 \\
& \space     & 79.42      & 75.61       & 74.64        & 79.34      & 79.42         \\
& \multirow{2}{*}{Student}  & ShuffleNet-V1 & ShuffleNet-V1 & MobileNet-V2  & MobileNet-V2 & ShuffleNet-V2  \\
& \space     & 70.50      & 70.50       & 64.60        & 64.60      & 71.82         \\ 
\midrule
\multirow{5}{*}{Features}
& FitNet~\cite{fitnets} & 73.59      & 73.73       & 64.14        & 63.16      & 73.54    \\
& RKD~\cite{rkd}        & 72.28      & 72.21       & 64.52        & 64.43      & 73.21    \\
& CRD~\cite{crd}        & 75.11      & 76.05       & 69.73        & 69.11      & 75.65    \\
& OFD~\cite{ofd}        & 75.98      & 75.85       & 69.48        & 69.04      & 76.82    \\
& ReviewKD~\cite{revkd} & \textbf{77.45}      & 77.14       & 70.37        & 69.89      & \textbf{77.78}    \\     
\midrule
\multirow{4}{*}{Logits}                                                         
& DKD~\cite{dkd}     & 76.45               & 76.70                        & 69.71          & 70.35              & 77.07       \\
& KD~\cite{kd}        & 74.07               & 74.83                        & 67.37          & 67.35              & 74.45         \\
& IKL-KD~\cite{cui2025decoupled}    &76.64  &77.19 &70.40 &70.62 &77.16 \\
& \textbf{GKL-KD}    &76.76      & \textbf{77.42}      & \textbf{70.61}    & \textbf{70.78}   & 77.49 \\
\bottomrule
\end{tabular}
}
\label{tab:cifar2_kd}
\end{table*}

\begin{table*}[h]
\centering
\small
\caption{\textbf{Comparisons with strong training settings on ImageNet for knowledge distillation.}}
\label{tab:comparision_strong}
{
\begin{tabular}{ccccc}
\toprule
 Method & KD & DKD &DIST &GKL-KD \\
\midrule
  Top-1 Accuracy (\%) & 80.89 & 80.77 &80.70 & \textbf{80.98} \\
\bottomrule
\end{tabular}
}
\end{table*}

\subsection{Comparisons on CIFAR-100 for Knowledge Distillation}
We experiment on CIFAR-100 with the following cases: 1) the teacher and student models have the same unit network architectures; 2) the teacher and student models have different unit network architectures. The results are listed in Table~\ref{tab:cifar_kd} and Table~\ref{tab:cifar2_kd}. We have achieved the best results in 4 out of 6 and 3 out of 5 experimental settings respectively.

Moreover, we follow the concurrent work~\cite{hao2023vanillakd} and conduct experiments with BEiT-Large as the teacher and ResNet-50 as the student under a strong training scheme, the experimental results are summarized in Table~\ref{tab:comparision_strong}. The model trained by GKL-KD shows slightly better results.

\subsection{More Ablation Studies}
\label{sec:appendix_ablation}
\minisection{Ablation on $\gamma$ for Adversarial Robustness} 
We have added a new sensitivity study where $\gamma$ is varied under the adversarial training task on balanced CIFAR-100. An interesting phenomenon is observed: First, a smoother class-wise weighting function $\varphi(\cdot)$ generally leads to higher clean accuracy, as reflected by the performance gain, confirming the rationale of introducing global information.
Second, as $\gamma$ increases with class-wise weighting function $\varphi(\cdot)$, adversarial robustness first improves
and then declines. This suggests that moderate values of
$\gamma$ provides a better balance between clean accuracy and
robustness, whereas overly large $\gamma$ may harm robustness due to the accompanying drop in clean accuracy.

\begin{table}[h]
    \centering
    \begin{tabular}{ccc}
         \toprule
         $\gamma$ & Clean &AA  \\
         \midrule
         -(KL baseline) &62.87 &30.29 \\
         \midrule
         \multicolumn{3}{c}{Sample-wise $\varphi(x_{m},x_{n})$} \\
         \midrule
         $\gamma$=0.0 &67.99 &30.98 \\
         $\gamma$=0.4 &64.15 &31.70 \\
         \midrule
         \multicolumn{3}{c}{Class-wise $\varphi(x_{m},x_{n})$} \\
         \midrule
         $\gamma$=0.0 &67.99 &30.98 \\
         $\gamma$=0.2 &66.69 &31.23 \\
         $\gamma$=0.3 &65.80 &31.59 \\
         $\gamma$=0.4 &65.76 &31.91 \\
         $\gamma$=0.5 &63.57 &31.80 \\
         \bottomrule
    \end{tabular}
    \caption{Ablation of $\gamma$ on balanced data CIFAR-100 for adversarial robustness.}
    \label{tab:gamma_adv_full}
\end{table}

\minisection{Ablation on Temperature ($\tau$) for Global Information}
As discussed in Sec.~\ref{sec: IKL},  the incorporated class-wise global information is proposed to promote intra-class consistency and mitigate the biases from sample noises.
When calculating the $\bar w_{y}$ and $\bar s_{y}$, a temperature $\tau$ could be applied before getting sample probability vectors.
We summarize the experimental results in Table~\ref{tab:ablation_tau} for ablation of $\tau$. Interestingly, we observe that models usually exhibit higher performance on clean images with a higher $\tau$. There are even 5.75\% improvements of clear accuracy while keeping comparable robustness when changing $\tau=1$ to $\tau=4$, which implies the vast importance of weights in $\mathbf{w}$MSE component of DKL/KL for adversarial robustness.
To achieve the strongest robustness, we finally choose $\tau=4$ as illustrated by empirical study. 
\begin{table}[h]
    \centering
   \begin{tabular}{ccc}
   \toprule
   \textbf{$\tau$} & Clean & AA \\
   \midrule
    $\tau=1$     & 59.99 &31.35 \\
    $\tau=2$     & 63.77 &31.88 \\
    $\tau=3$     & 65.28 &31.69 \\  
    $\tau=4$     & 65.76 &31.91 \\ 
    \bottomrule
    \end{tabular}
    \caption{Ablation on temperature $\tau$ for global information (Conference version paper for smoothing the weights).}
    \label{tab:ablation_tau}
\end{table}

\subsection{Other Applications with GKL}
\noindent{\bf Semisupervised learning.}
We use the open-sourced code from \url{https://github.com/microsoft/Semi-supervised-learning} and conduct semi-supervised experiments on CIFAR-100 with FixMatch and Mean-Teacher methods. Specifically, each class has 2 labeled images and 500 unlabeled images. All default training hyper-parameters are used for fair comparisons. We only replace the consistency loss with our GKL loss. As shown in Table~\ref{tab:dkl_semi}, with our GKL loss, the Mean-Teacher method even surpasses the FixMatch.

\begin{table*}[h]
\centering
\caption{\textbf{Semi-supervised Learning on CIFAR-100 with ViT-small backbone.}}
{
\begin{tabular}{cccc}
\toprule
 Method  & Pseudo-label &Consistency Loss   &Last epoch Top-1 Acc(\%) \\
\midrule
\multicolumn{4}{c}{\textbf{FixMatch}} \\
\midrule
FixMatch       &hard  &Cross-entropy Loss    &69.20 \\
\midrule
FixMatch       &soft  &Cross-entroy/KL Loss  &69.09 \\
FixMatch       &soft  &GKL Loss              &\textbf{70.00} \\
 \midrule
 \multicolumn{4}{c}{\textbf{Mean-Teacher}} \\
 \midrule
Mean-Teacher  &soft  &MSE Loss               &67.38 \\
Mean-Teacher  &soft  &GKL Loss               &\textbf{70.05}      \\
\bottomrule
\end{tabular}
}
\label{tab:dkl_semi}
\end{table*}

\noindent{\bf Semantic segmentation distillation.} We conduct ablation on the semantic segmentation distillation task. We use the APD~\cite{tian2022adaptive} as our baseline for their open-sourced code. All default hyper-parameters are adopted. We only replace the original KL loss with our GKL loss. As shown in Table~\ref{tab:dkl_seg}, we achieve better performance with the GKL loss function, demonstrating that the GKL loss can be complementary to other techniques in semantic segmentation distillation.

\begin{table*}[h]
\centering
\caption{\textbf{Semantic segmentation distillation with APD on ADE20K.}}
{
\begin{tabular}{ccccc}
\toprule
 Method  &Teacher &Student & Teacher mIoU   &Student mIoU \\
\midrule
Baseline                 &-          &ResNet-18 &-     &37.19  \\              
\midrule
APD with KL loss  &ResNet-101 &ResNet-18 &43.44 &39.25 \\
APD with GKL loss &ResNet-101 &ResNet-18 &43.44 &\textbf{39.75} \\
\bottomrule
\end{tabular}
}
\label{tab:dkl_seg}
\end{table*}

\subsection{Complexity of GKL}
Compared with the KL divergence loss, GKL loss is required to update the global class-wise prediction scores $W \in \mathbb{R}^{C \times C}$ where $C$ is the number of classes during training. 
This extra computational cost can be nearly ignored when compared with the model forward and backward. Algorithm~\ref{algo:algorithm_dkl} shows the implementation of our GKL loss in Pytorch style. On dense prediction tasks like semantic segmentation, $\Delta_{a}$ and $\Delta_{b}$ can require large GPU memory. Here, we also provide the memory-efficient implementations for $w$MSE loss component, which is listed in Algorithm~\ref{algo:algorithm_wmse}.

\begin{algorithm}
    \caption{Pseudo code for DKL/GKL loss in Pytorch style.}
    \begin{algorithmic}
        \State \textbf{Input:} $logits_{a}, logits_{b} \in \mathbb{R}^{B \times C}$, one-hot label $Y$, $W \in \mathbb{R}^{C \times C}$, $\alpha$, $\beta$, $\gamma$. 

        \State class\_scores = one-hot @ W; 
        \State class\_scores = torch.pow(class\_scores, $\gamma$);
        \State Sample\_weights = class\_scores.view(-1, C, 1) @ class\_scores.view(-1, 1, C); 
        \State $\Delta_a$ = $logits_a$.view(-1, C, 1) - $logits_a$.view(-1, 1, C);
        \State $\Delta_b$ = $logits_b$.view(-1, C, 1) - $logits_b$.view(-1, 1, C);
        \State wMSE\_loss = (torch.pow($\Delta_{n}$ - $\Delta_{a}$, 2) * Sample\_weights).sum()/ Sample\_weights.sum() * $\frac{1}{4}$;
        
        \State score\_a = F.softmax($logits_a$, dim=1).detach();
        \State log\_score\_b = F.log\_softmax($logits_b$, dim=-1);
        \State CE\_loss = -(score\_a * log\_score\_b).sum(1).mean(); 
        \State $\textbf{return}$ $\beta$ * CE\_loss + $\alpha$ * wMSE\_loss.
    \end{algorithmic}
\label{algo:algorithm_dkl}
\end{algorithm}

\begin{algorithm}
    \caption{Memory efficient implementation for wMSE\_loss in Pytorch style.}
    \begin{algorithmic}
        \State \textbf{Input:} $logits_{a}, logits_{b} \in \mathbb{R}^{B \times C}$, one-hot label $Y$, $W \in \mathbb{R}^{C \times C}$, $\gamma$;

        \State class\_scores = one-hot @ W; 
        \State class\_scores = torch.pow(class\_scores, $\gamma$);
        \State loss\_a = (class\_scores * $logits_a$ * $logits_a$).sum(dim=1) * 2 - torch.pow((class\_scores * $logits_a$).sum(dim=1), 2) * 2;
        \State loss\_b = (class\_scores * $logits_b$ * $logits_b$).sum(dim=1) * 2 - torch.pow((class\_scores * $logits_b$).sum(dim=1), 2) * 2;
        \State loss\_ex = (class\_scores * $logits_a$ * $logits_b$).sum(dim=1) * 4 - (class\_scores * $logits_a$).sum(dim=1) * (class\_scores * $logits_b$).sum(dim=1) * 4;
        \State wMSE\_loss = $\frac{1}{4}$ * (loss\_a + loss\_b - loss\_ex).sum() / torch.pow(class\_scores, 2).sum();
        \State $\textbf{return}$ wMSE\_loss.
    \end{algorithmic}
    \label{algo:algorithm_wmse}
\end{algorithm}

\subsection{\bf Connection between GKL and the Jensen-Shannon (JS) Divergence}
With the following JS divergence loss,
\begin{equation}
    JSD(P||Q) = \frac{1}{2} KL(P||M) + \frac{1}{2} KL(Q||M), \quad M=\frac{1}{2} P + \frac{1}{2} Q.
\end{equation}

We calculate its derivatives regarding $o_{n}$ (the student logits), 
\begin{eqnarray}
  \frac{\partial \mathcal{L}_{JSD}}{\partial \mathbf{o}_{n}^{i}} &=& \sum_{j=1}^{C}\mathbf{w}_{n}^{i,j} (\Delta \mathbf{n}_{i,j} - \Delta \mathbf{m'}_{i,j}) \label{eq:jsd_gradient}\\
  \textit{Softmax}(o_{m'}) &=& \frac{1}{2} s_{n} + \frac{1}{2} s_{m} \label{eq:jsd_constraint}
\end{eqnarray}
where $\mathbf{o}_{m}$ is the logits from the teacher model, $\mathbf{o}_{m'}$ is a virtual logits satisfying Eq.~\eqref{eq:jsd_constraint}, $\mathbf{s}_{m}=\textit{Softmax}(\mathbf{o}_{m})$, $\mathbf{s}_{n}=\textit{Softmax}(\mathbf{o}_{n})$, $\Delta \mathbf{m'}_{i,j}= \mathbf{o}_{m'}^{i} - \mathbf{o}_{m'}^{j}$, $\Delta \mathbf{n}_{i,j}= \mathbf{o}_{n}^{i} - \mathbf{o}_{n}^{j}$.

Correspondingly, the derivatives of GKL loss regrading $o_{n}$ (the student logits),
\begin{eqnarray}
    \frac{\partial \mathcal{L}_{GKL}}{\partial \mathbf{o}_{n}^{i}} = \underbrace{\alpha \sum_{j=1}^{C} \mathbf{w}_{m}^{i,j}(\Delta \mathbf{n}_{i,j} - \Delta \mathbf{m}_{i,j})}_{\textbf{Effects of wMSE}} + \underbrace{\beta * \mathbf{s}_{m}^{i} * (\mathbf{s}_{n}^{i} -1) + \mathbf{s}_{n}^{i} * (1-\mathbf{s}_{m}^{i})}_{\textbf{Effects of Cross-Entropy}}
\end{eqnarray}

Compared with GKL loss, the problem for JSD divergence in knowledge distillation is that:
\textit{The soft labels from the teacher models often embed dark knowledge and facilitate the optimization of the student models. However, there are no effects of the cross-entropy loss with the soft labels from the teacher model, which can be the underlying reason that JSD is worse than KD and GL-KD.}

As shown in Table~\ref{tab:jsd_vs_kl_dkl}, we also empirically demonstrate that GKL loss performs better than JSD divergence on the knowledge distillation task.

\begin{table*}[h]
\centering
\caption{\textbf{Comparisons between KL, GKL, and JSD on ImageNet-LT}.}
{
\begin{tabular}{ccccc}
\toprule
 Method &Student   &Teacher   &Teacher Acc(\%)  & Student Acc(\%) \\
\midrule
\multicolumn{5}{c}{\textbf{Self-distillation on Imbalanced Data}} \\
\midrule
 KL   &ResNet-50 &ResNet-50  &45.47  &47.04 \\
 JSD  &ResNet-50 &ResNet-50  &45.47  &46.64 \\
 Ours &ResNet-50 &ResNet-50  &45.47  &\textbf{47.50} \\
 \midrule
 \multicolumn{5}{c}{\textbf{Knowledge distillation on Imbalanced Data}} \\
 \midrule
 KL   &ResNet-50 &ResNeXt-101 &48.33 &48.31 \\
 JSD  &ResNet-50 &ResNeXt-101 &48.33 &47.82 \\
 Ours &ResNet-50 &ResNeXt-101 &48.33 & \textbf{49.40} \\
\bottomrule
\end{tabular}
}
\label{tab:jsd_vs_kl_dkl}
\end{table*}

\end{document}